\documentclass{llncs} 

\usepackage{graphicx}
\usepackage{subfig}
\graphicspath{{images/}{eps/}{tikz/}{./}}

\usepackage{amssymb}
\usepackage{amsmath}

\newcommand{\argmin}[1]{\underset{#1}{\operatorname{arg}\,\operatorname{min}}\;}

\usepackage[hidelinks]{hyperref}

\title{A 3D Face Modelling Approach for Pose-Invariant Face Recognition in a Human-Robot Environment}

\titlerunning{Pose-Invariant Face Recognition in a Human-Robot Environment}

\author{ Michael Grupp \inst{1}
\and
Philipp Kopp \inst{2}
\and
Patrik Huber \inst{3}
\and
Matthias R\"atsch \inst{2}
}

\institute{Technische Universit\"at M\"unchen, Munich, Germany\\
\email{michael.grupp@tum.de}
\and
Reutlingen University, Reutlingen, Germany\\
\email{philipp.kopp@student.reutlingen-university.de}, \email{matthias.raetsch@reutlingen-university.de}
\and
University of Surrey, Guildford, United Kingdom\\
\email{p.huber@surrey.ac.uk}
}

\begin{document}
\maketitle

\begin{abstract}
Face analysis techniques have become a crucial component of human-machine interaction in the fields of assistive and humanoid robotics. However, the variations in head-pose that arise naturally in these environments are still a great challenge.

In this paper, we present a real-time capable 3D face modelling framework for 2D in-the-wild images that is applicable for robotics. The fitting of the 3D Morphable Model is based exclusively on automatically detected landmarks. After fitting, the face can be corrected in pose and transformed back to a frontal 2D representation that is more suitable for face recognition. We conduct face recognition experiments with non-frontal images from the MUCT database and uncontrolled, in the wild images from the PaSC database, the most challenging face recognition database to date, showing an improved performance.

Finally, we present our SCITOS G5 robot system, which incorporates our framework as a means of image pre-processing for face analysis.
\end{abstract}


\section{Introduction}\label{sec:Intro}

Among the technologies for biometric identification and verification, face recognition has become a widely-used method, as it is a non-intrusive and reliable method.

However, the robustness of a facial recognition system is constrained by the degree of head pose that is involved and the recognition rates of state of the art systems drop significantly for large pose angles. Especially for tasks where a great variation in head pose has to be expected, like in human-robot interaction, pose-invariant face recognition is crucial. The rise of collaborative and assistive robots in industrial and home environments will increase the demand for algorithms that can adapt to changing settings and uncontrolled conditions. A humanoid robot's cognitive ability to interpret non-verbal communication during conversations relies also heavily on the face of the human counterpart.

The impact of head pose on the face analysis performance can be minimised by using normalisation techniques that transform non-frontal faces into frontal representations. This type of image pre-processing can loosely be classified into two categories:

\begin{itemize}
\item 2D methods (cylindrical warping, AAM (active appearance models)~\cite{cootesAAM}, 2D warping~\cite{paper:Asthana2D})
\item 3D methods (3DMM, GEM (generic elastic models)~\cite{prabhu_unconstrained_2011}, mean face shape~\cite{paper:Asthana3D})
\end{itemize}

As the 2D methods are trained on 2D data, the warping will only be an approximation of
the underlying 3D rotation. They also do not model imaging parameters like the pose of the face and illumination, and face parameters like expressions, explicitly. Instead, these parameters are inherent and implicitly encoded in the model parameters. When dealing with larger pose ranges than around $\pm$40$^\circ$ in yaw angle, a single 2D model is often no longer sufficient and a multi-model approach must be used (e.g. \cite{ASMcootes2004}, \cite{FengMultiModelApproach}).

The use of a 3D face model has several advantages, especially when further analysis involving the shape of the face is required, like in emotion detection. To correct the pose for face recognition, a 3D representation of the face can be rotated easily into a frontal view. At close ranges, the use of a 3D or RGB-D sensor can provide the additional depth information for the model. However, since the error of depth measurements increases quadratically with increasing distance when using RGB-D sensors like Microsoft Kinect~\cite{paper:KinectEval}, they are not suitable for acquiring image data from a distance. On the other hand, 2D cameras are more widely-used in existing platforms and cost-efficient, so that a method combining the advantages of 2D image acquisition and the analysis with a 3D model is needed.

The 3D Morphable Model (3DMM)  \cite{paper:3DMM_Vetter} satifies this need by providing a parameterised Principal Component Analysis (PCA) model for shape and albedo, which can be used for the synthesis of 3D faces from 2D images. 
A Morphable Model can be used to reconstruct a 3D representation from a 2D image through fitting. Fitting is the process of adapting the 3D Morphable Model in such a way that the difference to a 2D input image is as small as possible. Depending on the implementation, the fitter's cost function uses different model parameters to iteratively minimise the difference between the modeled image and the original input image. 

Existing fitting algorithms solve a complex cost function, using the image information to perform shape from shading, edge detection, and solve for both shape and texture coefficients (\cite{paper:Fitting_Romdhani,paper:Fitting_Rootseler,paper:Fitting_Tena,paper:Fitting_Hu,egger_pose_2014}). Due to the complexity of the cost functions, these algorithms require several minutes to fit a single image. While the execution time may not be crucial for offline applications, real-time robotics applications require faster solutions. More recently, new methods were introduced that use local features \cite{huber2015fitting} or geometric features (landmarks, edges) with a fitting algorithm similar to the iterative closest point algorithm \cite{bas2016fitting}.

In this paper, we present and evaluate a complete, fully automatic face modelling framework consisting of landmark detection, landmark-based 3D Morphable Model fitting, and pose-normalisation for face recognition purposes. The whole process is designed to be lightweight enough to run in real-time on a standard PC, but is also especially intended to be used for human-robot interaction in uncontrolled, in the wild environments. The theory is explained in Section \ref{sec:LM_3DMM}. In Section \ref{sec:Exp}, we show how the approach improves the recognition rates of a regular COTS (commercial off-the-shelf) face recognition system when it is used as a pre-processing method for pose-normalisation on non-frontal and uncontrolled images from two image databases. Finally, we show how our approach is suitable for real-time robotics applications by presenting its integration into the HMI framework of our SCITOS robot platform in Section \ref{sec:app}. Section \ref{sec:Conclusion} concludes the paper and gives an outlook to future work.

All parts of the framework are publicly available to support other researchers: the C++ implementations of the landmark detection \footnote{\url{https://github.com/patrikhuber/superviseddescent}} and the fitting algorithm and the face model \footnote{\url{https://github.com/patrikhuber/eos}}, as well as a demo app that combines both \footnote{\url{http://www.4dface.org}}. 


\section{Landmark-based 3D Morphable Model Fitting}\label{sec:LM_3DMM}
In this section, we will give a brief introduction to the 3D Morphable Model and then introduce our algorithm to recover pose and shape from a given 2D image. We will then present our pose-invariant texture representation in form of a so-called isomap.

\subsection{The 3D Morphable Model}

A 3D Morphable Model consists of a shape and albedo (colour) PCA model constructed from 169 3D scans of real faces. These meshes first have to be brought in dense correspondence, that is, vertices with the same index in the mesh correspond to the same semantic point on each face. The model used for this implementation consists of 3448 vertices. A 3D shape is expressed in the form of $\mathbf{v}= [x_1, y_1, z_1, \dots, x_{_V}, y_{_V}, z_{_V}]^\mathrm{T}$, where $[x_{v}, y_{v}, z_{v}]^\mathrm{T}$ are the coordinates of the $v$th vertex and $V$ is the number of mesh vertices. The RGB colour values are stacked in a similar manner. PCA is then applied to both the vertex- and colour data matrices separately, each consisting of $m$ stacked 3D face meshes, resulting in $m-1$ shape eigenvectors $\mathbf{S}_i$, their variances $\sigma_{S,i}^2$, and a mean shape $\mathbf{\bar{s}}$, and similarly for the colour model ($\mathbf{T}_i$, $\sigma_{T,i}^2$ and $\mathbf{\bar{t}}$). A face can then be approximated as a linear combination of the respective basis vectors:

\begin{equation}
\label{eq:pca_face}
\mathbf{s} = \mathbf{\bar{s}} + \sum_{i=1}^{m-1}\alpha_i \mathbf{S}_i, ~~~~
\mathbf{t} = \mathbf{\bar{t}} + \sum_{i=1}^{m-1}\beta_i \mathbf{T}_i,
\end{equation}

where $\boldsymbol{\alpha} = [\alpha_1, \dots ,\alpha_{m-1}]^\mathrm{T}$ and $\boldsymbol{\beta} = [\beta_1, \dots ,\beta_{m-1}]^\mathrm{T}$ are vectors of shape and colour coefficients respectively.

\subsection{Shape and Pose Reconstruction}

Given an image with a face, we would like fit the 3D Morphable Model to that face and obtain a pose invariant representation of the subject. At the core of our fitting algorithm is an affine camera model, shape reconstruction from landmarks, and a pose invariant textural representation of the face. We thus only use the shape PCA model from the 3DMM and not the colour PCA model, and instead use the original texture from the image to obtain the best possible image quality for subsequent face analysis steps.

Similar to Aldrian and Smith \cite{paper:Fitting_Aldrian}, we obtain a linear solution by decomposing the problem into two steps which can be alternated. The first step in our framework is to estimate the pose of the face. Given a set of 2D landmark locations and their known correspondences in the 3D Morphable Model, we compute an affine camera matrix. The detected 2D landmarks $ x_i \in \mathbb{R}^3 $ and the corresponding 3D model points $ X_i \in \mathbb{R}^4 $ (both represented as homogeneous coordinates) are normalised by similarity transforms that translate the centroid of the image and model points to the origin and scale them so that the Root-Mean-Square distance from their origin is $\sqrt{2}$ for the landmark and $\sqrt{3}$ for the model points respectively: $ \tilde{x}_i = \text{U}x_i $ with $ \text{U} \in \mathbb{R}^{3\times4}$, and $ \tilde{X}_i = \text{W}X_i $ with $ \text{W} \in \mathbb{R}^{4\times4} $. Using $ \geq 4 $ landmark points, we then compute a normalised camera matrix $ \tilde{\text{C}} $ using the \textit{Gold Standard Algorithm} \cite{Hartley2004} and obtain the final camera matrix after denormalising: $ \text{C} = \text{U}^{-1}\tilde{\text{C}}\text{W} $.

The second step in our framework consists of reconstructing the 3D shape using the estimated camera matrix. We estimate the vector of PCA shape coefficients $\boldsymbol{\alpha}_s$:
\begin{equation}
	\argmin{\boldsymbol{\alpha}_s}\sum_{i=1}^{3N} (x_{m,i} - x_{i})^{2} + \lambda \lVert \boldsymbol{\alpha}_s \rVert^{2} \,,
\end{equation}
where $N$ is the number of landmarks and $\lambda$ is a weighting parameter for the regularisation that is needed to only allow plausible shapes. $ x_i $ are the detected landmark locations and $ x_{m,i} $ is the projection of the 3D Morphable Model shape to 2D using the estimated camera matrix. Subsequently, the camera matrix can then be re-estimated using the now obtained identity specific face shape, instead of only the mean face. Both steps are iterated for a few times - each of them only involves solving a small linear system of equations.

In the experiments with automatically detected landmarks, we use a cascaded regression based approach similar to Feng et al.~\cite{paper:LandmarkSurrey}. After an initial estimate, i.e. placing the landmarks in the region found by a face detector, we extract local features (HOG \cite{dalal2005hog}) to update the landmark locations until convergence towards their actual position. The detected 2D landmarks have known corresponding points in the 3D Morphable Model, and these points are then used to fit the model. The combination of the regression based landmark detection and the landmark-based 3D Morphable Model fitting results in a lightweight framework that is feasible to run on videos (all components run in the order of milliseconds on a standard CPU).

\subsection{Pose-independent Face Representation}

After fitting the pose and shape, a correspondence between the 3D mesh and the face in the 2D image is known for each point in the image. We use this correspondence to remap the original face texture from the image onto a pose-invariant 2D surface that covers the face from all angles. We create such a generic representation with the isomap algorithm ~\cite{paper:TenenbaumIsomap}: it finds a projection from the 3D vertices to a 2D plane that preserves the geodesic distance between the mesh vertices.
Our mapping is computed with the algorithm from Tena~\cite{ThesisTena}.

The isomap of different persons are in dense correspondence with each other, meaning each location in the map corresponds to the same physical point in the face of every subject (for example, a hypothetical point $ x = [100, 120] $ is always the center of the right eye). It can consequently show the accuracy of the fitting and is therefore a plausible representation of the result. 

\begin{figure}[!ht]
\centering
\includegraphics[width=2.5in]{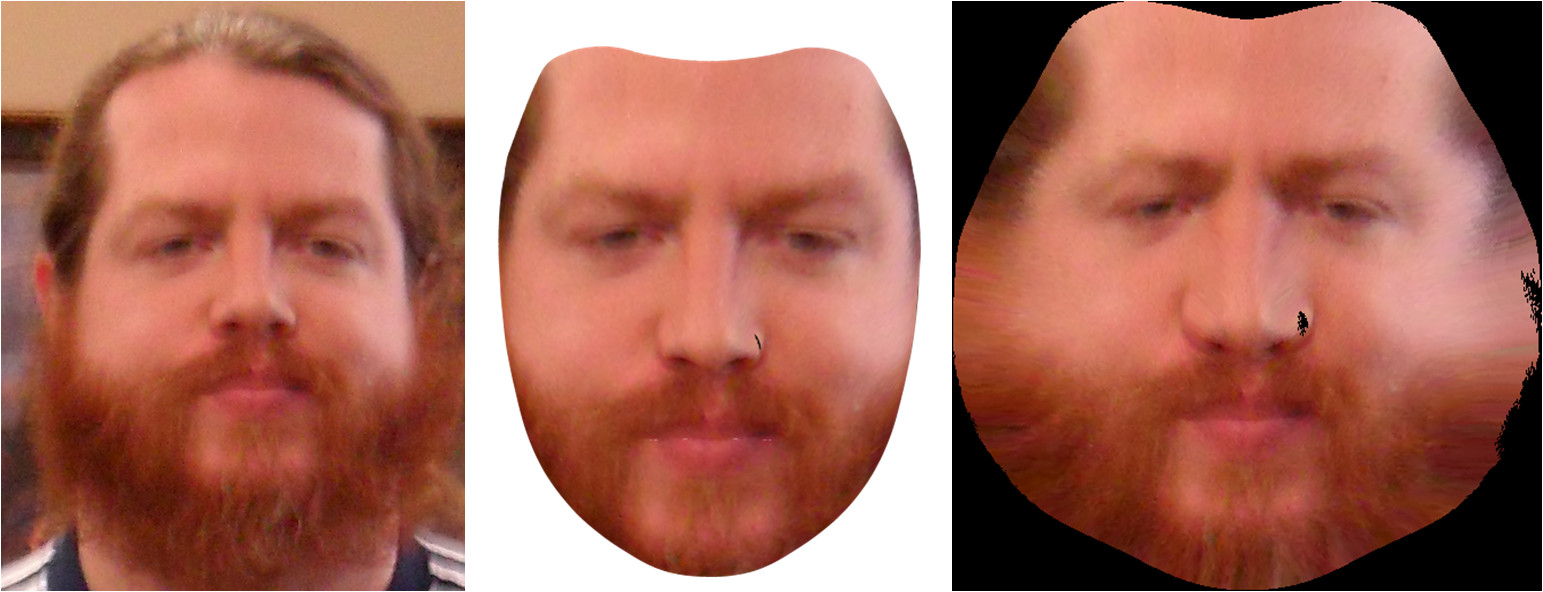}
\caption{Example picture from PaSC (left), frontal rendering (middle) and pose invariant isomap representation (right).}
\label{fig_isomap}
\end{figure}

Figure~\ref{fig_isomap} shows an isomap and a frontal rendering side by side. The isomap captures the whole face, while the rendering only shows the frontal parts of the face. As the face in the input image is frontal, with most parts of the face visible, there is little self occlusion. In case of large poses, a major part of the isomap can not be filled with face texture, and these regions will be marked black (for example, a small black spot next to the nose can be observed in the figure).

\section{Face Recognition Experiments} \label{sec:Exp}

To analyse the performance of our approach, we have run experiments on two different image databases (MUCT and Point and Shoot Challenge (PaSC)). Our experiments are face verification experiments, which means that the system has to verify the identity of probe images compared to a set of gallery images via matching. The results are then statistically analysed regarding the ratio of the two possible error types, the false acceptance rate (FAR) and false rejection rate (FRR) and plotted into detection error tradeoff (DET) curves \cite{paper:DET}, using the ground-truth subject-ID information provided in the metadata of the image databases. In the experiments on the PaSC database (section \ref{sec:PaSC}), we filter the probe and gallery images according to different head poses.

We used a market-leading, commercial face recognition engine for the process of enrolment and matching\footnote{Our results obtained with commercial systems should not be construed as the vendors' full-capability results.}. As a reference measure, all experiments were also conducted using the original, unprocessed images from the databases (denoted as \textit{org}). With this reference, we are able to analyse the impact of using our approach compared to a conventional face recognition framework without 3D modelling.

\subsection{Experiments on the MUCT Image Database} \label{sec:MUCT}
The MUCT Database consists of 3755 faces of 276 subjects and is published by the University Of Cape Town \cite{paper:MUCT}. For our experiments only persons without glasses and their mouths closed were used. After this filtering, 1221 pictures were left. The pictures are taken from 5 different cameras that cause different pose angles and there are in total 10 different lighting schemes applied to the pictures. Moreover, the size of the faces within the images is quite large and the subjects are placed in a controlled lab environment. MUCT comes with 76 manually labeled landmarks given for every image, from which we used 16 for the initialisation of our landmark-based 3DMM fitting (\textit{LM}). 

\begin{figure}[h!]
\centering
	\includegraphics[width=0.5\columnwidth]{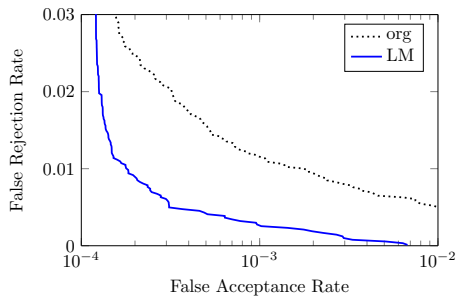}
\caption{DET curve showing the fewer recognition errors when using our landmark-based fitting method (\textit{LM}) on the MUCT image database. The dotted line shows the result for the experiment with the original images without pose normalisation (\textit{org}).}
\label{fig_MUCT}
\end{figure}

In total, we used the unprocessed images (\textit{org}) and the pose normalised renderings coming from our fitting method (\textit{LM}). We ran a verification experiment on these images, leading to similarity matrices with 1~490~841 scores for both methods. Using these scores, a DET curve was plotted, of which a relevant part is depicted in Figure \ref{fig_MUCT}. 

It is clear to see that the verification errors of the face recognition drop significantly when our landmark-based modelling approach is used to correct the head pose. This experiment shows that when pose is the major uncontrolled factor, like on MUCT, the usage of our \textit{LM} aproach leads to a clearly improved face recognition performance. 

\subsection{Experiments on the Point and Shoot (PaSC) Image Database}\label{sec:PaSC}

For our proposed use case in assistive or humanoid robotics, MUCT is not representative of typical in the wild images we encounter. To verify the performance in heavily uncontrolled settings, we decided to do further investigations on a larger, more challenging database. The Point and Shoot Challenge (PaSC) is a performance evaluation challenge for developers of face recognition systems initiated by the Colorado State University \cite{paper:PaSC}.

As the name depicts, the images and videos used in the challenge are taken with consumer point and shoot cameras and not with the help of professional equipment. The database offers 9376 still images of varying resolution up to 4000$\times$3000 pixels and labeled metadata. The arrangement of the image sceneries raises some major challenges for the automated recognition of faces. The pictures were taken in various locations, indoors and outdoors, with complex backgrounds and harsh lighting conditions, accompanied by a low image quality due to blur, noise or incorrect white balance. Additionally, the regions of interest for face recognition are also relatively small and of low resolution because the photographed people are not the main subject of the scenery.

In a direct comparison, the verification rate on PaSC equals only 0.22 at 0.001 FAR, in contrast to the more controlled databases MBE (0.997), GBU (0.8) and LFW (0.54) \cite{paper:PaSC}. 

Because of these harsh, uncontrolled conditions, which are similar to a real life scenario, we chose this image database for our experiments and applied our fitting algorithm (\textit{LM}) to the images. Like the previous experiment on MUCT (sec. \ref{sec:MUCT}), it was a verification experiment using the same COTS face recognition algorithm.

\subsubsection{Influence of the Number of Landmarks}

\begin{figure*}[ht]
\centering
	\subfloat[]{\raisebox{0.8cm}{\includegraphics[width=0.2\textwidth]{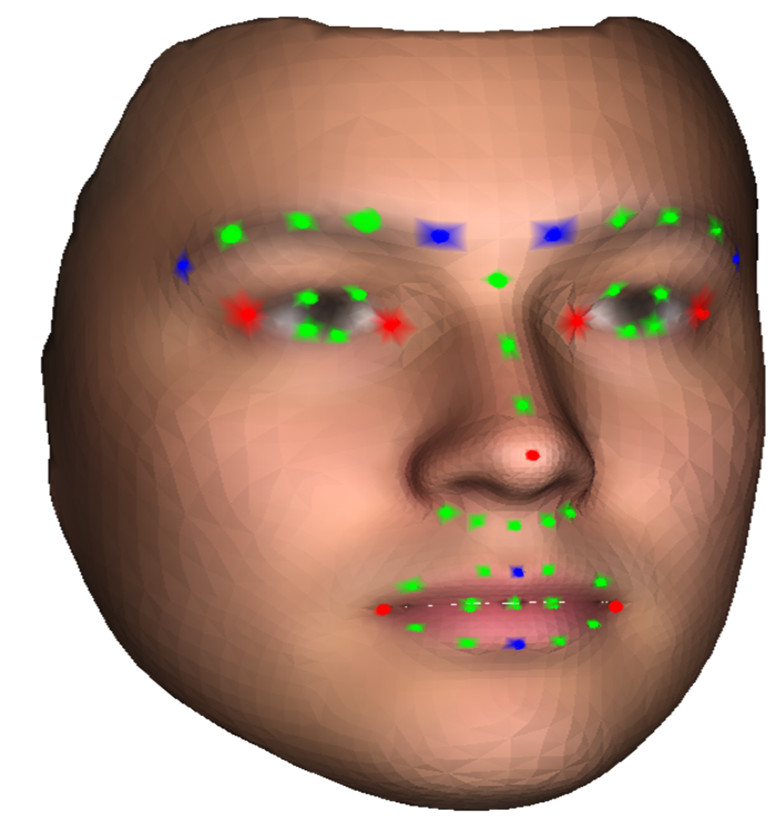}}} \hspace{1.5cm}
	\subfloat[]{\includegraphics[width=0.65\textwidth]{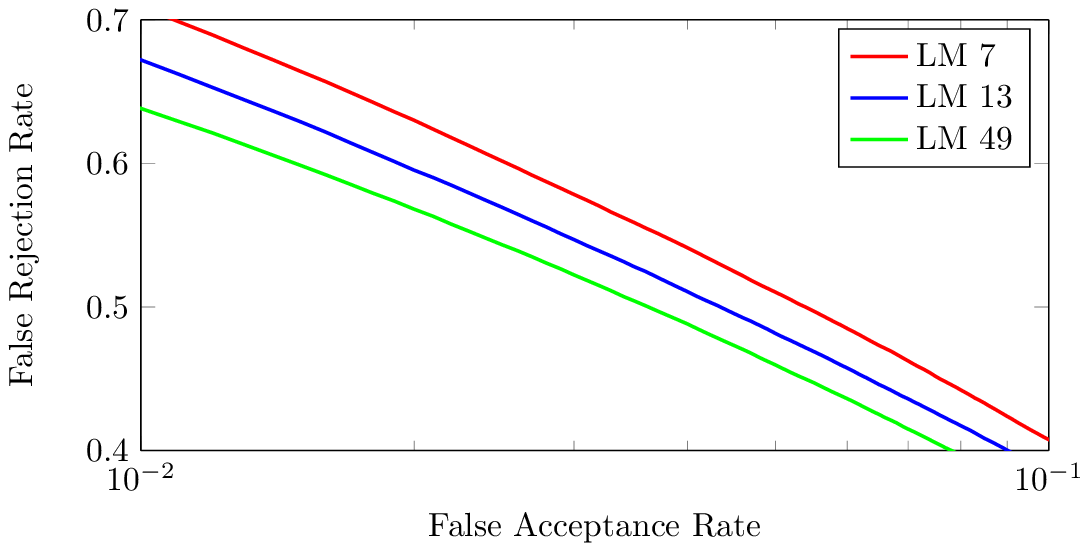}}
\caption{Evaluation of different sets of facial landmarks. (a) visualises the landmark sets on the mean face of the 3D Morphable Model. red: 7; red+blue: 13; red+blue+green: 49. (b) shows a DET curve for face recognition on PaSC when using the different sets of landmarks (7, 13 and 49) for our \textit{LM} fitting algorithm.}
\label{fig_DET_landmarks}
\end{figure*}

Our experiments on PaSC are intended to be as near to a realistic scenario as possible. Therefore, we used the automatic landmark detection of \cite{paper:LandmarkSurrey} instead of using manually annotated facial landmarks. The detector and our fitting algorithm (\textit{LM}) allow the use of different landmark schemes, which can vary in their amount. We tested three different sets of 7, 13 and 49 landmarks, which are visualised on the 3D Morphable Model in Figure \ref{fig_DET_landmarks} (a). To compare the different amounts of landmarks, their corresponding results for the face recognition experiment on PaSC are shown in Figure \ref{fig_DET_landmarks} (b). It can be observed that a higher number of facial landmarks leads to a higher quality 3DMM fitting, which then leads to a better verification performance.

\subsubsection{Performance Evaluation across Head Pose}

One of the key questions of this work is whether our approach can improve the performance of face recognition under uncontrolled conditions, especially when dealing with head pose. 

To allow a more detailed interpretation of the results, we generated head pose annotations for the PaSC database. We used OpenCV's implementation of the POSIT algorithm \cite{paper:POSIT} to calculate the yaw angles for each image in the database. POSIT estimates the pose of an object from an image using a 3D reference model and corresponding feature points of the object in the image. In our case, the 3D model was the 3D Morphable Model and the feature points were facial landmarks.

\begin{figure}
    \centering
    \includegraphics[width=0.4\columnwidth]{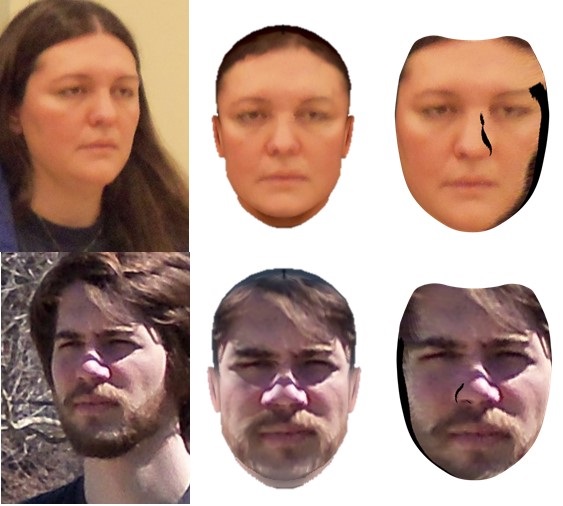}
    \caption{From left to right: example images from PaSC (cropped to area of face for this figure) and their frontalised renderings from the commercial face modelling solution and our landmark-based fitter.}
    \label{fig:CE_LM}
\end{figure}

We used this yaw angle annotation to filter the scores into different groups of yaw angles. The results obtained with our method were then compared to the results without any face modelling. For comparison, we also tested the performance of a commercial face modelling system (\textit{CFM}) using the same workflow. 
The \textit{CFM} uses additional modelling capabilities like mirroring (to reconstruct occluded areas of the face) and deillumination. Figure \ref{fig:CE_LM} depicts frontalised renderings of the commercial system and our fitter. In total, we compared all images of PaSC against each other and the yaw angle filter separated the scores into 10$^{\circ}$ bins that span from -70$^{\circ}$ to +70$^{\circ}$. 

\begin{figure}[ht!]
\centering
	\includegraphics[width=0.5\columnwidth]{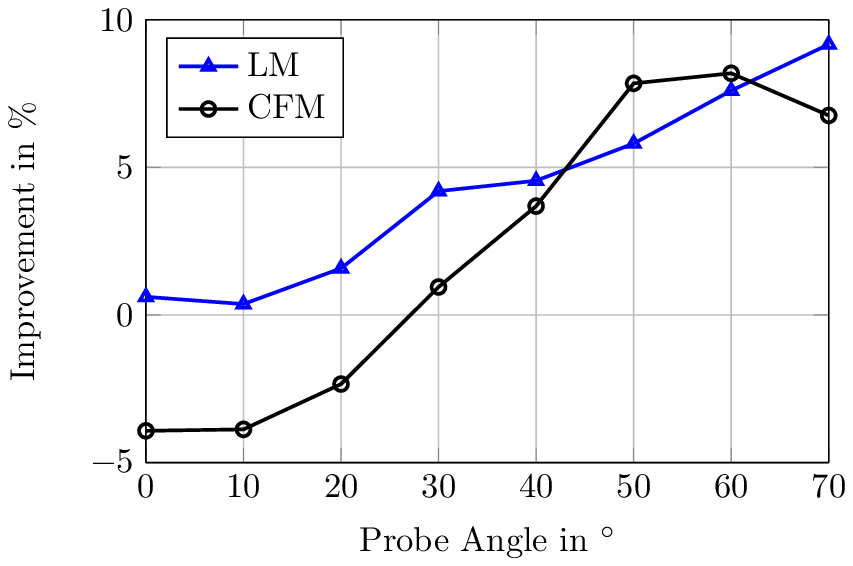}
\caption{Face recognition improvement over the baseline COTS system of two face modelling approaches. Different probe angles are compared against a frontal gallery. \emph{(blue):} our 3DMM landmark fitting (\textit{LM}), \emph{(black):} commercial face modelling system (\textit{CFM}).}
\label{fig_curveDeltaYaw_0}
\end{figure}

\begin{figure*}[!ht]
\centering
	\includegraphics[width=0.5\columnwidth]{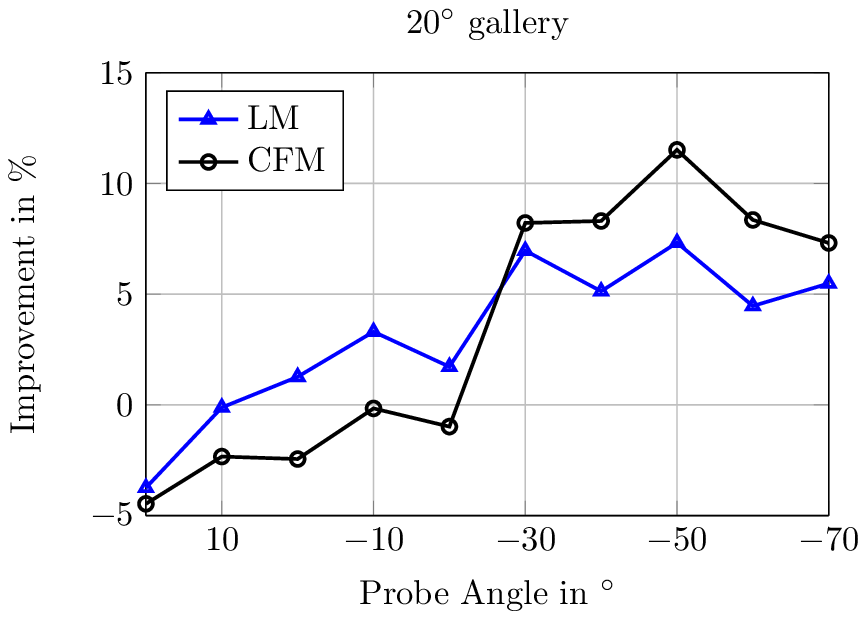}
	\includegraphics[width=0.5\columnwidth]{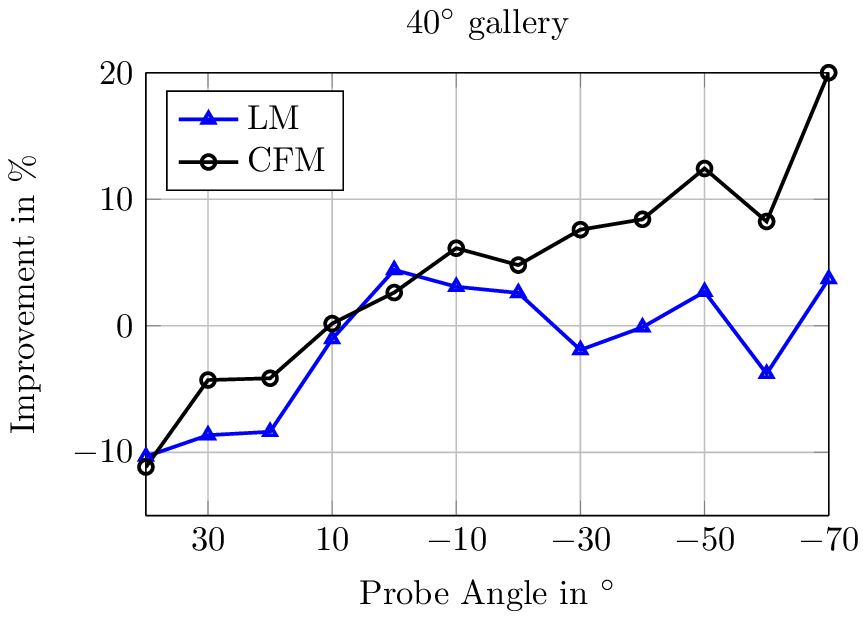}
\caption{Face recognition improvement over the baseline COTS system of two face modelling approaches. Different probe angles are compared against a 20$^\circ$ and 40$^\circ$ gallery. \emph{(blue):} our 3DMM landmark fitting (\textit{LM}), \emph{(black):} commercial face modelling system (\textit{CFM}).}
\label{fig_curveDeltaYaw_20_40}
\end{figure*}

To better see how the pose normalisation methods improve the performance, the resulting curves show the absolute difference of the false rejection rate to the baseline without face modelling (\textit{org}) at a specific operating point. This type of diagram allows a more simple and clearly arranged representation of the results at different head poses.
 
In Figure \ref{fig_curveDeltaYaw_0}, such a curve from the experiment is shown. A set of gallery images of 0$^{\circ}$ yaw is compared to sets of probe images from 0$^{\circ}$ to 70$^{\circ}$. As gallery databases in face recognition applications often contain frontal images, this experiment reflects a common use case. At the operating point 0.01 FAR, we then plot the improvement for each group of yaw angles. The commercial face modelling solution shows its first improvements over the conventional face recognition starting at 30$^{\circ}$ yaw. Surprisingly, our 3D Morphable Model based method is even able to slightly improve the matching performance for frontal images. Although the COTS face recognition system was used outside of the specification for the large yaw angles, we are able to improve the performance for the whole pose range. By using the original image's texture, our fitter (\textit{LM}) does not alter the face characteristics and corrects only rotations in the pitch or roll angles.

In the next two experiments, we take a look at how well the system operates when the gallery is also not frontal. In Figure \ref{fig_curveDeltaYaw_20_40}, the results for the experiment with a 20$^{\circ}$ and a 40$^{\circ}$ gallery can be seen. In this case, matching the faces is a much more difficult task, especially when the subjects look in different directions and large areas of the faces are not visible due to self-occlusion. While we are still able to improve the face recognition capabilities in the experiment with a 20$^{\circ}$ gallery, our method \textit{LM} struggles to keep up with the commercial solution for a 40$^{\circ}$ gallery. The ability to reconstruct parts of the non-visible texture is a key component of the commercial modelling system (\textit{CFM}) in such a use case which our system currently doesn't offer. Although, for the robotics application, which we present in section \ref{sec:app} of this paper, it is likely that frontal images are used for the gallery enrolment.


\section{Application on a Robot System}\label{sec:app}

Our fast and accurate 3D face modelling method is particularly suitable for face analysis tasks in assistive or humanoid robotics. In the following, we demonstrate our efforts in this field by illustrating how we integrate our approach into a mobile robot's software framework.

\begin{figure*}[!t]
\centering
\includegraphics[width=1\textwidth]{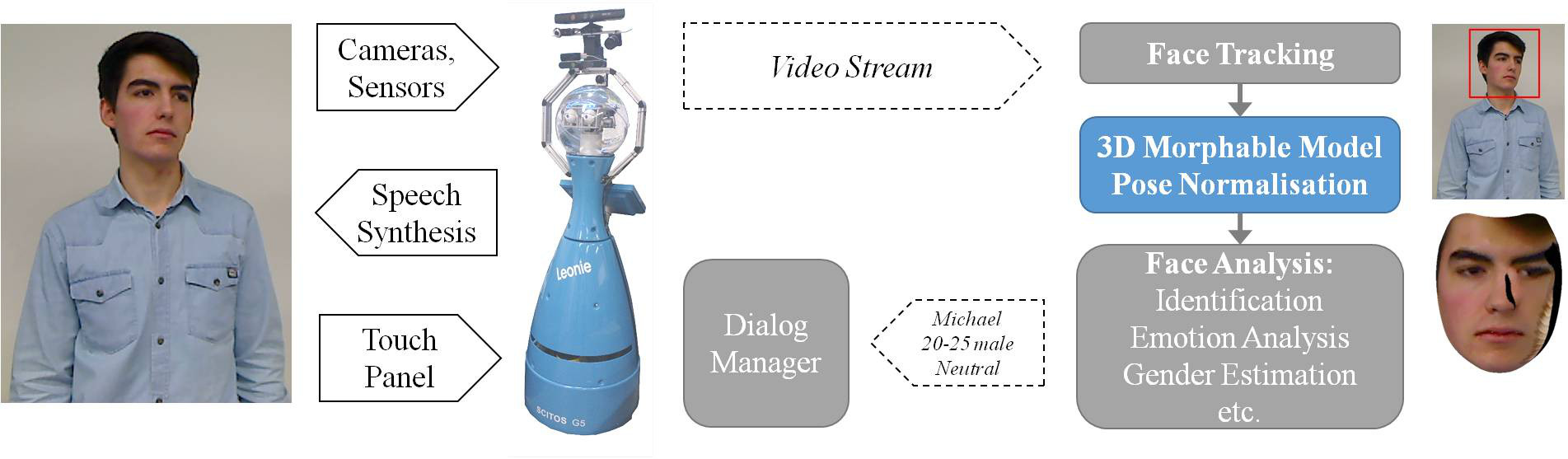}
\caption{Schematic overview of the framework for human-robot interaction on our SCITOS G5 platform. Our proposed landmark based 3D Morphable Model fitter acts as a pre-processing method for various face analysis tasks.}
\label{fig_Leonie}
\end{figure*}

A mobile robot platform combined with an HMI system is the basis for our research on human-robot interaction. The robot is based on a SCITOS G5 drive system with onboard industrial PC which is able to navigate autonomously in indoor buildings. For mapping and localisation ability, as well as collision avoidance, two laser scanners are attached to the base (SICK S300 and Hokuyo URG-04LX). Using this base platform, the robot is able to approach humans safely. The HMI part of the robot consists of a touch-based display and a head with moveable, human-like eyes. On top of the robot, we mounted a pan-tilt-unit (PTU; Directed Perception D46-17), where additional sensors and cameras can be installed. The PTU adds an additional degree of freedom which is used for camera positioning to enhance the tracking process.

To align the cameras to the region of interest, we track people by applying an adaptive approach which uses a particle filter to track the position and size of the target and estimates the target motion using an optical flow based prediction model. Furthermore, an SVM-based learning strategy is implemented to calculate the particle weights and to prevent
bad updates while staying adaptive to changes in object pose \cite{paper:adaptiveTracking,paper:InfoInside2015}. At this point, the 3DMM pose normalisation of our approach can be used for the image of the tracked face. It can be either applied continuously or if the pose of the subject is greater than a certain threshold (using the knowledge obtained in the experiments of section \ref{sec:PaSC}).

\begin{figure*}[!t]
\centering
\includegraphics[width=0.9\textwidth]{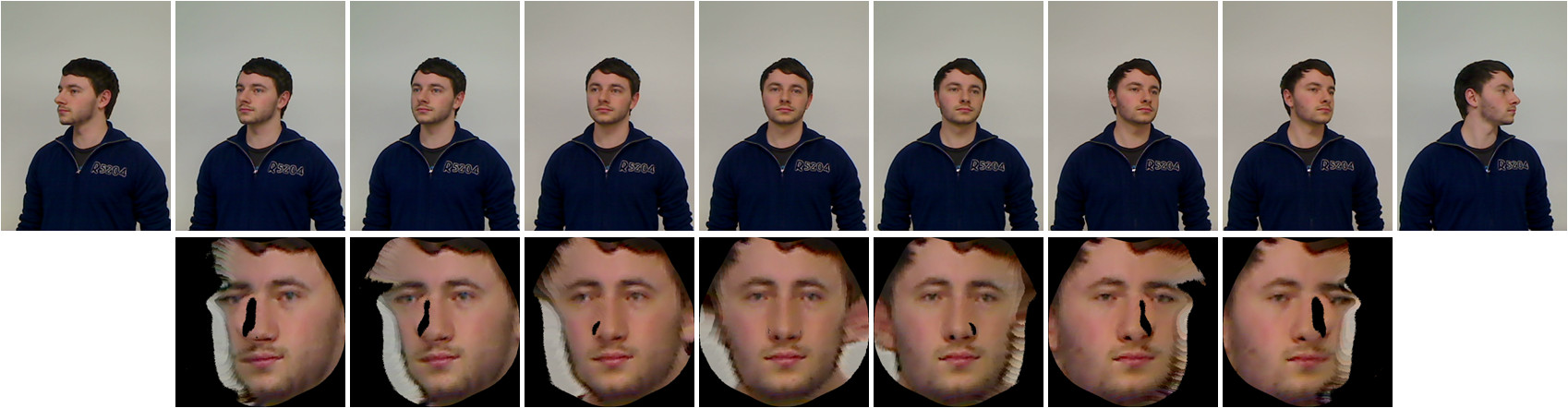}
\caption{Frames from a video of a subject turning from left profile to right profile view. The bottom row shows the pose-invariant isomap representations after fitting. Black regions depict self-occlusion/invisible parts of the face and thus cannot be drawn. On the two most extreme pose images, the initialisation with the face detector fails.}
\label{fig_Filmstreifen_Philipp}
\end{figure*}

Key frames of an example video taken from our robot system are presented in the upper row of Figure \ref{fig_Filmstreifen_Philipp}. The OpenCV object detection with a face cascade \cite{paper:ViolaJones} was used to initialise the regression based landmark detector, which then automatically found 15 facial landmarks. The fitting results, represented in isomaps, are shown in the second row of Figure \ref{fig_Filmstreifen_Philipp}.
On the robot, we use the frontal renderings as an input for the same face recognition engine that was also used for the experiments. To distinguish the person's identity properly, the system should generate a rather high score for this positive match. The pose normalised renderings allow the face recognition algorithm to achieve a clearly higher score compared to the original, unprocessed images. Possible use cases for a robot of this kind are industrial mobile robotics, shop or museum tour guides or a surveillance system (e.g. in a supermarket). In all these cases, the robot has to interact naturally with technically unskilled people. It is therefore convenient to be able to analyse people independent of their pose, without requiring them to look at the robot directly. Identification, age and gender estimation on pose normalised renderings offer the possibility to personalise the robot’s behaviour.


\section{Conclusion and Further Work}\label{sec:Conclusion}

We proposed a powerful method for head pose correction using a fully automatic landmark-based approach for fitting a 3D Morphable Model. By using an efficient fitting algorithm, our approach can be used for tasks which require a fast real-time solution that can be also used on live video streams. In contrast to existing work, we focus on an evaluation of our approach with automatically found landmarks and in-the-wild images and publish the whole 3DMM fitting pipeline as open source software (see section \ref{sec:Intro}).

An experimental evaluation on the two commonly-used image databases MUCT and PaSC showed the significance of our approach as a means of image processing for facial recognition in both controlled and heavily unconstrained settings. For recognition algorithms that are mostly trained for frontal faces, a rotation of the head means a loss of information that results in a lower matching score. We showed that our methodology is capable of improving the recognition rates for larger variations in head pose. Compared to a commercial face modelling solution, our approach keeps up well and even outperforms it in certain scenarios. The addition of 3DMM pose normalisation certainly brings advantages compared to a conventional face recognition framework when the setting is uncontrolled - like in robotics, surveillance or consumer electronics.

Furthermore, we presented our software framework on a mobile robot platform which uses our pose normalisation approach to allow a more reliable face analysis. We are currently doing further research on age- and emotion-estimation systems to expand the abilities of our robot. In the future, we also plan to statistically evaluate the effect on their performance when using our landmark-based pose normalisation, like we did for face recognition in this paper.


\section*{Acknowledgements}
The authors would like to thank Huan Fui Lee and the RT-LIONS robocup team of Reutlingen University. We would also like to thank CyberExtruder, Inc. for supporting our research.


\clearpage


\end{document}